\documentclass{article} 
\usepackage{iclr2024_conference,times}

\usepackage{amsmath,amsfonts,bm}

\def\eqref#1{equation~\ref{#1}}

\def\1{\bm{1}}
\newcommand{\train}{\mathcal{D}}

\DeclareMathAlphabet{\mathsfit}{\encodingdefault}{\sfdefault}{m}{sl}
\SetMathAlphabet{\mathsfit}{bold}{\encodingdefault}{\sfdefault}{bx}{n}

\def\gD{{\mathcal{D}}}

\def\gH{{\mathcal{H}}}

\def\gL{{\mathcal{L}}}

\def\gO{{\mathcal{O}}}

\newcommand{\E}{\mathbb{E}}

\usepackage{hyperref}
\usepackage{url}
\usepackage{microtype}
\usepackage{xcolor}
\usepackage{adjustbox}
\usepackage{wrapfig}
\usepackage{graphicx,subcaption}
\usepackage{booktabs} 
\usepackage{placeins}

\title{In-Context Data Distillation with TabPFN}

\author{Junwei Ma, Valentin Thomas, Guangwei Yu, Anthony Caterini \\
Layer 6 AI\\
\texttt{\{jeremy,valentin.t,guang,anthony\}@layer6.ai}
}

\newcommand{\aref}[1]{\hyperref[#1]{appendix~\ref*{#1}}}

\usepackage{amsmath}
\usepackage{amssymb}
\usepackage{mathtools}
\usepackage{amsthm}

\usepackage[capitalize,noabbrev]{cleveref}
\theoremstyle{plain}

\theoremstyle{definition}

\theoremstyle{remark}

\usepackage[textsize=tiny]{todonotes}

\newcommand{\distill}{\mathcal{D}_{\text{dist}}}
\renewcommand{\train}{\mathcal{D}_{\text{train}}}

\newcommand{\distillloss}{\gL(\train \rightarrow \distill) }

\iclrfinalcopy 
\begin{document}

\maketitle

\begin{abstract}
Foundation models have revolutionized tasks in computer vision and natural language processing. However, in the realm of tabular data, tree-based models like XGBoost continue to dominate. TabPFN, a transformer model tailored for tabular data, mirrors recent foundation models in its exceptional in-context learning capability, being competitive with XGBoost's performance without the need for task-specific training or hyperparameter tuning. Despite its promise, TabPFN's applicability is hindered by its data size constraint, limiting its use in real-world scenarios. To address this, we present in-context data distillation (ICD), a novel methodology that effectively eliminates these constraints by optimizing TabPFN's context. ICD efficiently enables TabPFN to handle significantly larger datasets with a fixed memory budget, improving TabPFN's quadratic memory complexity but at the cost of a linear number of tuning steps. Notably, TabPFN, enhanced with ICD, demonstrates very strong performance against established tree-based models and modern deep learning methods on 48 large tabular datasets from OpenML.
\end{abstract}

\section{Introduction}
\label{introduction}

\looseness=-1 Recent advancements in foundation models \citep{bommasani2021opportunities}, including GPT-3 \citep{brown2020language}, CLIP \citep{radford2021learning}, and diffusion models \citep{ho2020denoising}, have demonstrated remarkable proficiency in tasks related to computer vision and natural language processing, marking a significant leap forward in the field of machine learning. However, despite its pivotal role across diverse industries \citep{tang2020customer, benjelloun2020google, sattarov2023findiff}, tabular data has long been dominated by tree-based models like XGBoost \citep{chen2016xgboost}. The predominance of these models is attributed to several factors. Firstly, unlike their deep learning counterparts, tree-based models require minimal hyper-parameter tuning, simplifying their use and boosting their adoption. Secondly, while deep learning models owe their success to pre-training and transfer learning, this advantage is largely absent in tabular data due to its diverse nature \citep{borisov2022deep} and lack of pre-training datasets such as ImageNet \citep{deng2009imagenet}. 

Recently, the introduction of TabPFN \citep{hollmann2023tabpfn} and its variants~\citep{ma2023tabpfgen,dooley2023forecastpfn} have marked a significant advance in this domain. Trained on a vast array of synthetic datasets, TabPFN harnesses the transfer learning capabilities of deep learning models while matching the performance of tree-based models on smaller datasets, without the need for extensive hyper-parameter tuning. However, due to its reliance on the Transformer~\citep{vaswani2017attention} architecture, it suffers from quadratic memory complexity scaling, which renders it impractical for large real-world datasets.

\begin{figure}
\centering
\begin{subfigure}[t]{.24\linewidth}
\centering\includegraphics[width=\linewidth, trim={0 0 0 7mm},clip]{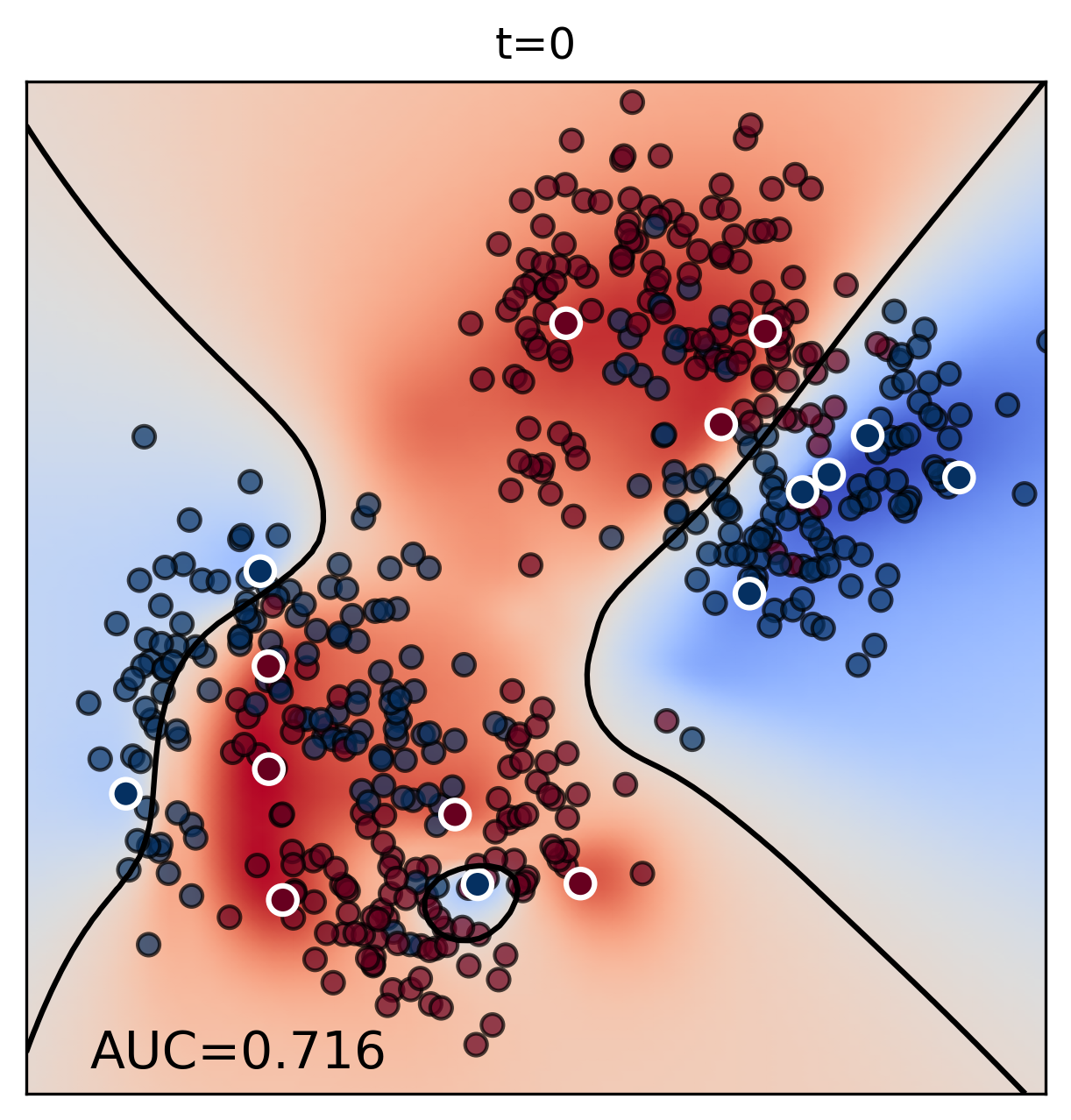}
\caption{$t=0$}
\end{subfigure}
\begin{subfigure}[t]{.24\linewidth}
\centering\includegraphics[width=\linewidth, trim={0 0 0 7mm},clip]{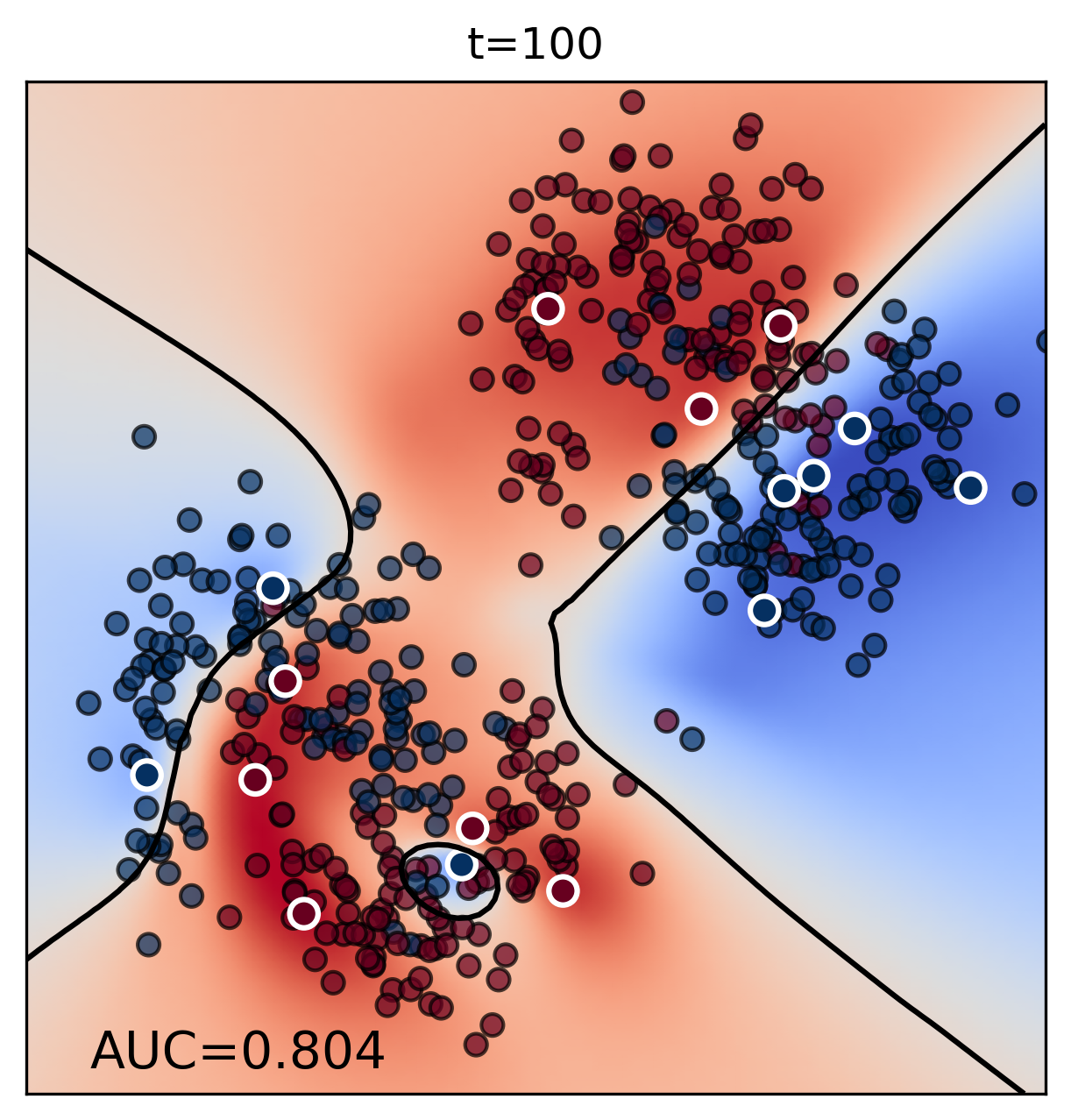}
\caption{$t=100$}
\end{subfigure}
\begin{subfigure}[t]{.24\linewidth}
\centering\includegraphics[width=\linewidth, trim={0 0 0 7mm},clip]{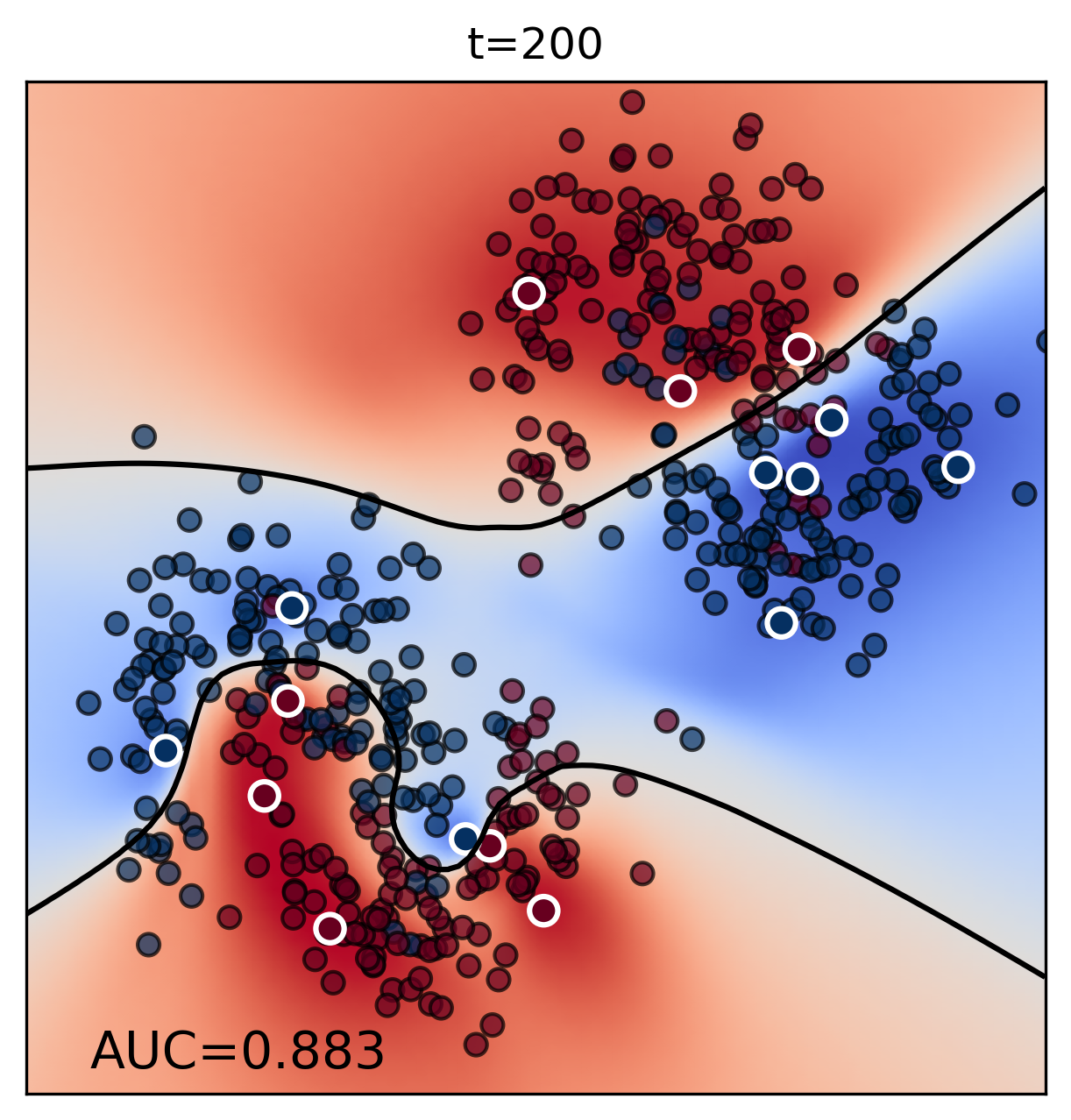}
\caption{$t=200$}
\end{subfigure}
\begin{subfigure}[t]{.24\linewidth}
\centering\includegraphics[width=\linewidth, trim={0 0 0 7mm},clip]{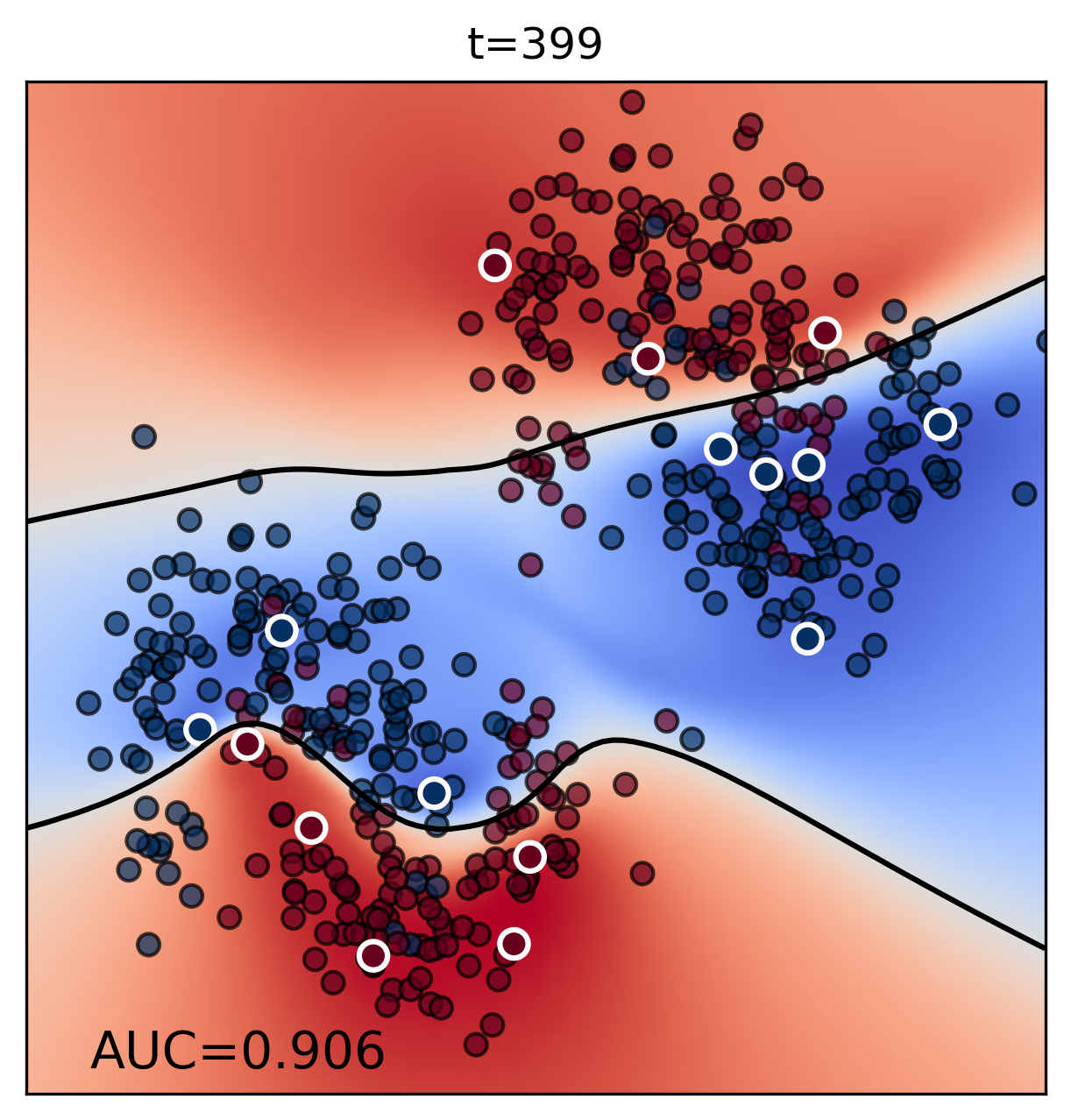}
\caption{$t=400$}
\end{subfigure}
\caption{\looseness=-1 Evolution of the distilled datapoints (circles with white borders, only 8 per class) and our decision boundary on a simple two double moons 2d dataset. The black line represents the decision boundary $p(y|x, \distill)=0.5$ of TabPFN conditioned on the 16 distilled points. The shaded red (respectively blue) regions represent which class the classifier will assign to datapoints in that region. The circles with a black edge color are the test dataset. The distilled points are initialized on random training points and over time they move around in order to improve the classifier.}
\label{fig:distill_example}
\end{figure}

 The concept of prompt tuning, prevalent in the LLM literature~\citep{lester2021power, li2021prefixtuning}, posits that adapting a pre-trained model through prompt modification, rather than adjusting the model weights, can be more effective. This is particularly relevant for TabPFN, whose strength lies in in-context meta-learning across a multitude of datasets. As the context used in TabPFN is the training set, this method can be also be interpreted through the lens of dataset distillation~\citep{wang2018dataset}. As shown in \autoref{fig:distill_example}, our method is able to optimize the distilled datapoints in order to refine TabPFN's decision boundaries and improve its classification accuracy.

Building on this idea, we introduce \emph{In-Context Distillation} (ICD). This novel approach performs backpropagation on the context tokens, allowing us to extend the original TabPFN to larger datasets. Moreover, TabPFN with ICD demonstrates superior performance over both the original TabPFN and other competitive models across 48 large OpenML benchmark datasets \citep{bischl2021openml, mcelfresh2023neural}.

\section{Background and related work}
\label{section:background}

\textbf{TabPFN}~\citep{hollmann2023tabpfn} is a novel transformer-based model developed for classifying tabular data, differing from traditional methods~\citep{houthooft2016vime,somepalli2021saint,arik2021tabnet} that require a unique model for each task. Instead, TabPFN adopts a different approach, commonly found in vision and text foundation models, by training on a diverse array of tasks~\citep{muller2022transformers} with a shared set of weight parameters $\theta$ applicable across all tasks. In order to perform classification on a new dataset, one uses the training data as \textit{context} and the new data to classify as a \textit{query}. While this allows making predictions extremely fast, only at the cost of inference, there are inherent limits to the approach. TabPFN suffers from the $\gO(n^2)$ memory scaling of transformers, restricting its applicability to only datasets with less than 1000 datapoints and 100 features. Attempts at context selection through sketching methods have yet to yield success \citep{feuer2023scaling}.

\textbf{Dataset distillation} is research area pioneered by~\citet{wang2018dataset}, aiming at learning a small set of training examples -- called the \textit{distilled set} $\distill$ -- such that a classifier trained on only those few examples can achieve a performance comparable to a classifier trained on the original dataset $\train$. While there have been advances simplifying the approach~\citep{zhao2021dataset}, the main strategy is to optimize $\distill$ directly by backpropagation through the entire neural network training procedure.

\textbf{Prompt-Tuning} \citep{lester2021power, li2021prefixtuning} has recently emerged as a new field in machine learning, which specifically addresses efficient fine-tuning of large foundational models. Unlike other parameter-efficient fine-tuning methods \citep{xu2023parameter}, prompt-tuning significantly reduces the number of trainable parameters. This efficiency becomes particularly apparent in models like TabPFN as we will see in \autoref{section:in-context-distillation}. Instead of using prompts as context inputs, TabPFN uses the real data. Consequently, prompt-tuning in the context of TabPFN can also be seen as data distillation, where the optimization is done on the input data itself, rather than a pure prompt-based approach.

\section{In-Context Distillation}
\label{section:in-context-distillation}
As mentioned in \autoref{section:background}, TabPFN is limited to small tabular datasets. This is because the training examples are used as \textit{context} to classify new examples. It can be natural to consider data distillation as a way to resolve that issue by compressing a large dataset into one that can fit in the context of TabPFN.
However, naively applying data distillation techniques such as \citet{wang2018dataset} would require partially retraining the whole TabPFN model several times, which can be very expensive. In this paper, we propose a simpler alternative: we view the distilled dataset as the context, which is learned to maximize the likelihood of the training data. 

More precisely, we consider the following setting: we have access to a classifier denoted by $p_\theta$ trained on a variety of datasets. Given a new dataset $\train$ and a query point $x$ to classify, its class probability is computed as $p_\theta(y|x, \train)$, where $\train$ can be understood as the ``prompt''. 
In-Context Distillation (ICD) -- similarly to prompt-tuning~\citep{lester2021power} -- optimizes the likelihood of the real data given the distilled one:
$$\distillloss \triangleq -\E_{(x,y) \sim \train} [\log p_\theta(y|x, \distill)].$$
By minimizing $\distillloss$, we obtain a dataset of size $|\distill| \ll |\train|$ such that the training data has a high likelihood. Then, given a new point $x$ to classify, we use $\distill$ as the context and predict the label $\arg \max_y p_\theta(y|x, \distill)$. While for small datasets we could directly compute $\arg \max_y p_\theta(y|x, \train)$, for larger datasets $\train$ is too large to be used as context.

While this objective function $\distillloss$ is similar to previous works in dataset distillation, we can directly compute the gradient $\nabla_{\distill} \distillloss$ without having to retrain $\theta$ because TabPFN's inference is conceptually similar to the whole training process of traditional methods.

Indeed, TabPFN is a foundation model for tabular data and it has been trained on a large number of tasks. As such, TabPFN is able to use the distilled data efficiently, even if it does not the follow the original data distribution. Moreover, traditional models that do not perform in-context learning have to be (at least partially) trained on the distilled data in order to generate useful gradients with respect to $\distill$.

\begin{figure}[t]
     \centering
     \begin{subfigure}[b]{0.64\linewidth} 
         \includegraphics[trim=0 0 0 20, width=\linewidth]{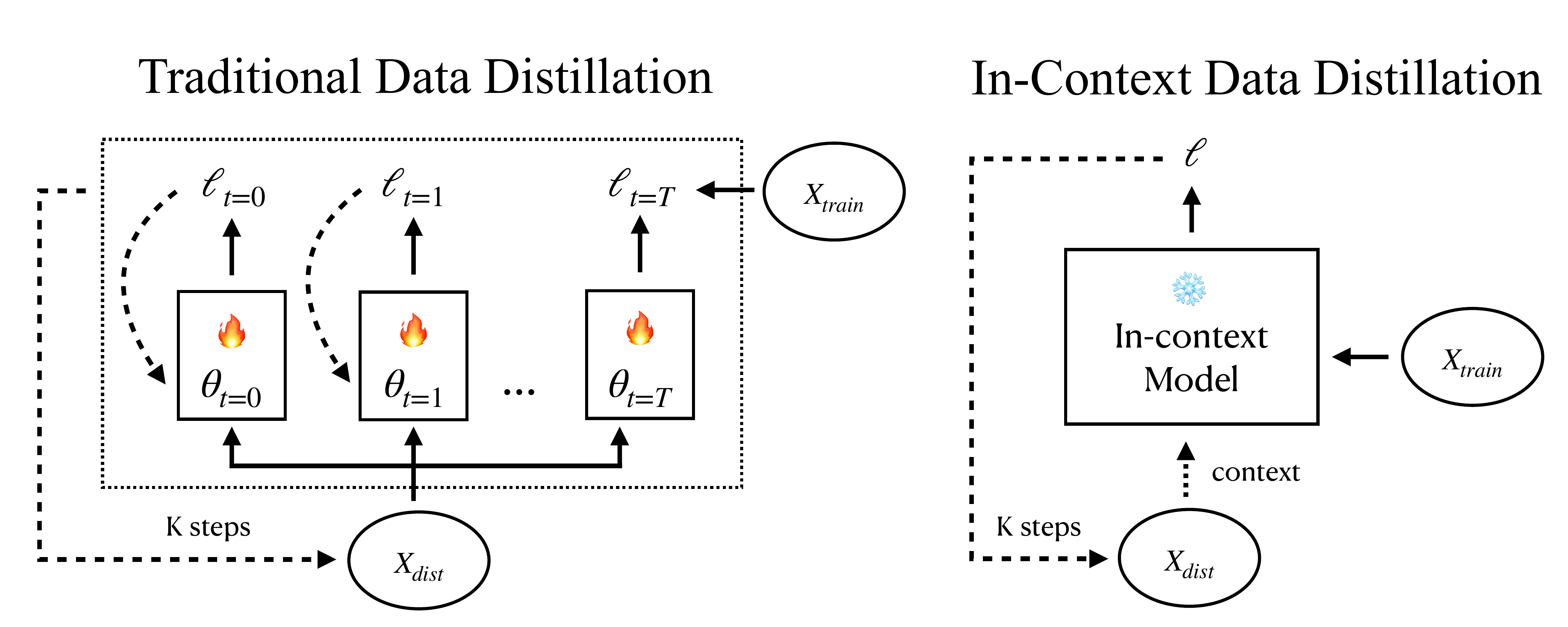}
         \caption{}
         \label{fig:distill-pfn-idea}
     \end{subfigure}
     \hfill 
     \begin{subfigure}[b]{0.31\linewidth} 
        \begin{adjustbox}{valign=t,raise=2.9cm}
         \includegraphics[trim=0 0 0 20, width=\linewidth]{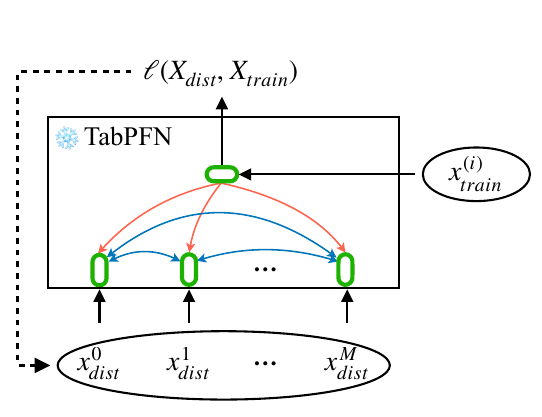}
         \end{adjustbox}
         \caption{}
         \label{fig:distill_main_figure}
     \end{subfigure}
     \label{fig:main_figure}
     \caption{(a) Comparison between traditional and in-context data distillation. \textbf{Left}: Traditional data distillation methods usually consist of nested optimization loops, where the inner loop optimizes model $\theta$ for $T$ steps, then an outer loop optimizes through the inner optimization process to update $X_{dist}$ for $K$ steps. \textbf{Right}: In-context data distillation only requires a single loop of optimization on $X_{dist}$ for $K$ steps, eliminating the need to optimize through the inner optimization process. (b) Detailed architecture diagram of applying ICD on TabPFN. Dashed lines indicate gradient flow. Blue and red arrows represent attentions within TabPFN. }
\end{figure}

\paragraph{Comparison with dataset distillation~\citep{wang2018dataset}}
In comparison, traditional models do not use a context. These do not generalize to new distributions or distribution shifts well and as such need to be trained for a number of steps on each data distribution. For these reasons, the distillation objective is of the form

\begin{align*}
    \arg \min_{\distill} \gL({\color{red}\distill}) &\triangleq -\E_{(x,y) \sim \gD} [\log(y|x, \theta_T({\color{red}\distill})], \\
    \text{s.t} \quad \theta_{t+1}({\color{red}\distill}) &= \theta_t({\color{red}\distill}) - \eta \nabla_\theta \E_{{(x,y) \sim \color{red}\distill}}\ell(y|x, \theta_t({\color{red}\distill})). \\
\end{align*}
As we can see, the dependence on $\distill$ is now implicit through $\theta$ and as such updating the distilled dataset for just one step requires backpropagation through $T$ steps of training as shown in \autoref{fig:distill-pfn-idea}. Note that several lines of work have proposed more efficient but approximate methods such as matching gradient statistics~\citep{zhao2021dataset}.

However, \textbf{distillation methods usually lower the performance of downstream models}. In contrast, because our motivation is to obtain the most information about $\train$ and fit it into the context, \textbf{our method actually increases the performance of the base model} compared to when it is trained on $|\distill|$ random points as is usually done~\citep{mcelfresh2023neural,feuer2023scaling}.

\section{Experiments}
\label{section:experiments}

        \paragraph{Experimental Setup}
\label{subsection:experimental_setup}
        Our experiments and analyses are based on the Tabzilla benchmark \citep{mcelfresh2023neural}, which comprises 176 datasets sourced from OpenML \citep{bischl2021openml}, a widely recognized repository for tabular data. We use 48 out of all the available datasets that have between $2{,}000$ to $200{,}000$ instances and do not contain NaNs, the specifics of which are detailed in \aref{section:dataset_details}. 
Each dataset is partitioned into train, validation, and test sets with a 80:10:10 split. For all methods, we used for early stopping on validation AUC to prevent overfitting. The performance of the model was evaluated based on the best-validated model's score on the test set, providing a reliable measure of its generalization capability. To ensure a fair and comprehensive comparison of the models, we report on three performance metrics: accuracy, F1 score, and Area Under the ROC Curve (AUC). In particular, AUC and F1 are known to give a more complete view of the performance of the model, especially in the case of imbalanced datasets (which most are, see \autoref{table:openml-datasets}).

\begin{wrapfigure}{R}{0.4\textwidth}
\vspace{-1em}
        \centering
        \includegraphics[width=0.38\textwidth]{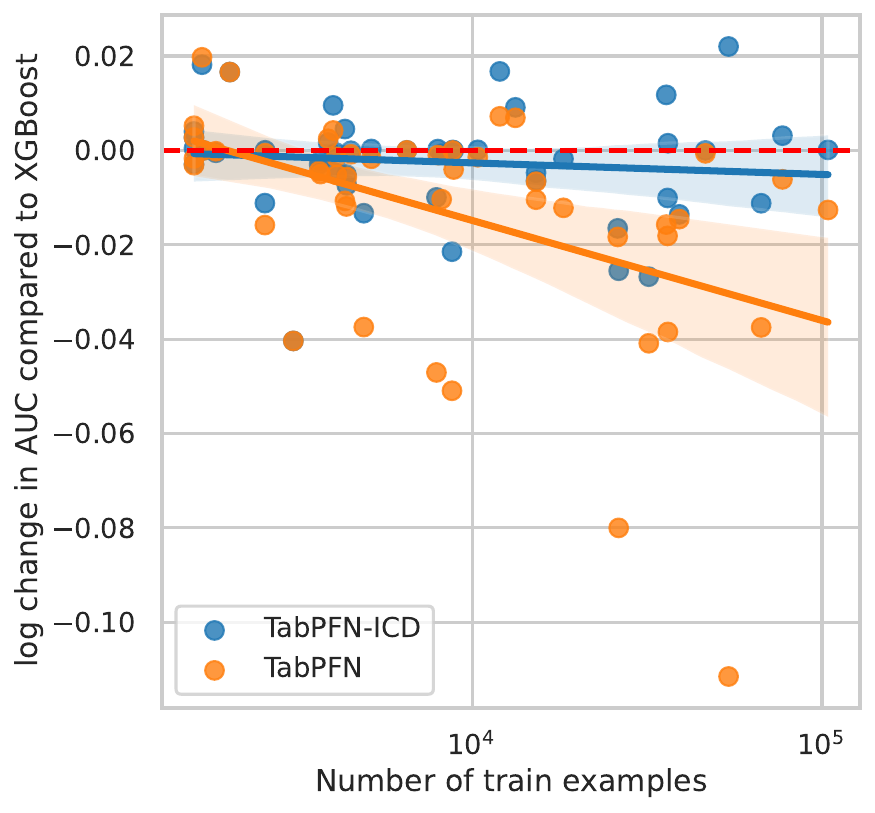}
        \caption{\looseness=-1 Log change in median AUC TabPFN-ICD and TabPFN have over XGB as a function of training set size.}
        \label{fig:auc_ratio_vs_train}
        \vspace{-3em}
\end{wrapfigure}

For \textbf{TabPFN-ICD} we use a constant learning rate of $10^{-2}$ with the Adam optimizer~\citep{kingma2015adam}. We always use $|\distill| = 1{,}000$. In this preliminary work, we only learn $X_\text{dist}$ and not $y_\text{dist}$. We choose to partition $\distill$ into sets of equal size for each class.
For TabPFN, we sample at random $1{,}000$ points in each dataset and use them for classification on new points.
\paragraph{Results}
\label{subsection:results}
\looseness=-1 We report the median of each metric aggregated over the 48 datasets in \autoref{table:median_table}. While currently our method does not beat a tuned XGBoost, its rank is always second in each metric and it significantly outperforms its base version TabPFN, and XGBoost with default hyperparameters. Furthermore, contrary to~\citet{mcelfresh2023neural}, we only used datasets with size larger than $2{,}000$ samples, resulting in TabPFN having significantly lower performance, even compared to XGBoost with default parameters.

In \autoref{fig:auc_ratio_vs_train}, we see that TabPFN's performance drops as a function of the training set size compared to XGBoost ($p_\text{value}<10^{-3}$ for rejecting the null hypothesis ``$\gH_0$: slope is 0''). This confirms our original motivation and echoes past work \citep{feuer2023scaling} that for many large datasets, subsampling a small subset of it is not enough to guarantee strong performance. With TabPFN-ICD, we observe that the trend in performance follows the one of XGBoost ($p_\text{value}\approx     0.45$), confirming that we are able to distill information about the dataset into a small number of samples. 

\begin{table}[]
    \centering
    \resizebox{\textwidth}{!}{
\begin{tabular}{lrrrrrrrrr}
\toprule
 & XGBoost (Tuned) & TabPFN-ICD & XGBoost & TabPFN & TabNet & RandomForest & SAINT & MLP & VIME \\
\midrule
Median AUC & \textbf{0.969} & \underline{0.967}& 0.953 & 0.951 & 0.939 & 0.928 & 0.878 & 0.864 & 0.824 \\
Median f1 & \textbf{0.921} & \underline{0.899} & 0.893 & 0.847 & 0.887 & 0.862 & 0.853 & 0.840 & 0.760 \\
Median accuracy & \textbf{0.923} & \underline{0.902} & 0.894 & 0.844 & 0.887 & 0.866 & 0.855 & 0.841 & 0.762 \\
\bottomrule
\end{tabular}
}
    \caption{Median accuracy, AUC, and F1 score for \textbf{TabPFN-ICD} and a variety of baselines calculated over the 48 datasets. Highest performance is bolded and the second-highest is underlined in each row.
    }
    \label{table:median_table}
\end{table}

\section{Discussion and conclusion}
We have proposed a straightforward yet innovative method for in-context distillation, drawing inspiration from both dataset distillation and prompt tuning techniques. This approach enables us to significantly enhance the scalability of TabPFN, allowing it to handle very large datasets while maintaining competitive performance against state-of-the-art algorithms. We believe that our method can open up new research avenues for data distillation and in-context learning for foundation models.

\bibliography{iclr2024_conference}

\begin{thebibliography}{27}
\providecommand{\natexlab}[1]{#1}
\providecommand{\url}[1]{\texttt{#1}}
\expandafter\ifx\csname urlstyle\endcsname\relax
  \providecommand{\doi}[1]{doi: #1}\else
  \providecommand{\doi}{doi: \begingroup \urlstyle{rm}\Url}\fi

\bibitem[Arik \& Pfister(2021)Arik and Pfister]{arik2021tabnet}
Sercan~{\"O} Arik and Tomas Pfister.
\newblock Tabnet: Attentive interpretable tabular learning.
\newblock In \emph{Proceedings of the AAAI conference on artificial intelligence}, volume~35, pp.\  6679--6687, 2021.

\bibitem[Benjelloun et~al.(2020)Benjelloun, Chen, and Noy]{benjelloun2020google}
Omar Benjelloun, Shiyu Chen, and Natasha Noy.
\newblock Google dataset search by the numbers.
\newblock In \emph{The Semantic Web -- ISWC 2020}, pp.\  667--682, 2020.

\bibitem[Bischl et~al.(2021)Bischl, Casalicchio, Feurer, Gijsbers, Hutter, Lang, Gomes~Mantovani, van Rijn, and Vanschoren]{bischl2021openml}
Bernd Bischl, Giuseppe Casalicchio, Matthias Feurer, Pieter Gijsbers, Frank Hutter, Michel Lang, Rafael Gomes~Mantovani, Jan van Rijn, and Joaquin Vanschoren.
\newblock {OpenML} benchmarking suites.
\newblock In \emph{Proceedings of the Neural Information Processing Systems Track on Datasets and Benchmarks}, 2021.

\bibitem[Bommasani et~al.(2021)Bommasani, Hudson, Adeli, Altman, Arora, von Arx, Bernstein, Bohg, Bosselut, Brunskill, et~al.]{bommasani2021opportunities}
Rishi Bommasani, Drew~A Hudson, Ehsan Adeli, Russ Altman, Simran Arora, Sydney von Arx, Michael~S Bernstein, Jeannette Bohg, Antoine Bosselut, Emma Brunskill, et~al.
\newblock On the opportunities and risks of foundation models.
\newblock \emph{arXiv preprint arXiv:2108.07258}, 2021.

\bibitem[Borisov et~al.(2022)Borisov, Leemann, Se{\ss}ler, Haug, Pawelczyk, and Kasneci]{borisov2022deep}
Vadim Borisov, Tobias Leemann, Kathrin Se{\ss}ler, Johannes Haug, Martin Pawelczyk, and Gjergji Kasneci.
\newblock Deep neural networks and tabular data: A survey.
\newblock \emph{IEEE Transactions on Neural Networks and Learning Systems}, 2022.

\bibitem[Brown et~al.(2020)Brown, Mann, Ryder, Subbiah, Kaplan, Dhariwal, Neelakantan, Shyam, Sastry, Askell, et~al.]{brown2020language}
Tom Brown, Benjamin Mann, Nick Ryder, Melanie Subbiah, Jared~D Kaplan, Prafulla Dhariwal, Arvind Neelakantan, Pranav Shyam, Girish Sastry, Amanda Askell, et~al.
\newblock Language models are few-shot learners.
\newblock \emph{Advances in neural information processing systems}, 33:\penalty0 1877--1901, 2020.

\bibitem[Chen \& Guestrin(2016)Chen and Guestrin]{chen2016xgboost}
Tianqi Chen and Carlos Guestrin.
\newblock {XGBoost}: A scalable tree boosting system.
\newblock In \emph{Proceedings of the 22nd ACM SigKDD International Conference on Knowledge Discovery and Data Mining}, pp.\  785--794, 2016.

\bibitem[Deng et~al.(2009)Deng, Dong, Socher, Li, Li, and Fei-Fei]{deng2009imagenet}
Jia Deng, Wei Dong, Richard Socher, Li-Jia Li, Kai Li, and Li~Fei-Fei.
\newblock Imagenet: A large-scale hierarchical image database.
\newblock In \emph{2009 IEEE Conference on Computer Vision and Pattern Recognition}, pp.\  248--255, 2009.
\newblock \doi{10.1109/CVPR.2009.5206848}.

\bibitem[Dooley et~al.(2023)Dooley, Khurana, Mohapatra, Naidu, and White]{dooley2023forecastpfn}
Samuel Dooley, Gurnoor~Singh Khurana, Chirag Mohapatra, Siddartha Naidu, and Colin White.
\newblock Forecastpfn: Synthetically-trained zero-shot forecasting.
\newblock \emph{arXiv preprint arXiv:2311.01933}, 2023.

\bibitem[Feuer et~al.(2023)Feuer, Hegde, and Cohen]{feuer2023scaling}
Benjamin Feuer, Chinmay Hegde, and Niv Cohen.
\newblock Scaling tabpfn: Sketching and feature selection for tabular prior-data fitted networks.
\newblock \emph{arXiv preprint arXiv: 2311.10609}, 2023.
\newblock URL \url{https://arxiv.org/abs/2311.10609v1}.

\bibitem[Ho et~al.(2020)Ho, Jain, and Abbeel]{ho2020denoising}
Jonathan Ho, Ajay Jain, and Pieter Abbeel.
\newblock Denoising diffusion probabilistic models.
\newblock \emph{Advances in neural information processing systems}, 33:\penalty0 6840--6851, 2020.

\bibitem[Hollmann et~al.(2023)Hollmann, M{\"u}ller, Eggensperger, and Hutter]{hollmann2023tabpfn}
Noah Hollmann, Samuel M{\"u}ller, Katharina Eggensperger, and Frank Hutter.
\newblock {TabPFN}: A transformer that solves small tabular classification problems in a second.
\newblock In \emph{International Conference on Learning Representations}, 2023.

\bibitem[Houthooft et~al.(2016)Houthooft, Chen, Duan, Schulman, De~Turck, and Abbeel]{houthooft2016vime}
Rein Houthooft, Xi~Chen, Yan Duan, John Schulman, Filip De~Turck, and Pieter Abbeel.
\newblock Vime: Variational information maximizing exploration.
\newblock \emph{Advances in neural information processing systems}, 29, 2016.

\bibitem[Kingma \& Ba(2015)Kingma and Ba]{kingma2015adam}
Diederik~P Kingma and Jimmy Ba.
\newblock Adam: A method for stochastic optimization.
\newblock In \emph{International Conference on Learning Representations}, 2015.

\bibitem[Lester et~al.(2021)Lester, Al-Rfou, and Constant]{lester2021power}
Brian Lester, Rami Al-Rfou, and Noah Constant.
\newblock The power of scale for parameter-efficient prompt tuning.
\newblock \emph{Conference on Empirical Methods in Natural Language Processing}, 2021.
\newblock \doi{10.18653/v1/2021.emnlp-main.243}.

\bibitem[Li \& Liang(2021)Li and Liang]{li2021prefixtuning}
Xiang~Lisa Li and Percy Liang.
\newblock Prefix-tuning: Optimizing continuous prompts for generation.
\newblock \emph{Annual Meeting of the Association for Computational Linguistics}, 2021.
\newblock \doi{10.18653/v1/2021.acl-long.353}.

\bibitem[Ma et~al.(2023)Ma, Dankar, Stein, Yu, and Caterini]{ma2023tabpfgen}
Junwei Ma, Apoorv Dankar, George Stein, Guangwei Yu, and Anthony Caterini.
\newblock Tabpfgen--tabular data generation with tabpfn.
\newblock In \emph{NeurIPS 2023 Second Table Representation Learning Workshop}, 2023.

\bibitem[McElfresh et~al.(2023)McElfresh, Khandagale, Valverde, Ramakrishnan, Goldblum, White, et~al.]{mcelfresh2023neural}
Duncan McElfresh, Sujay Khandagale, Jonathan Valverde, Ganesh Ramakrishnan, Micah Goldblum, Colin White, et~al.
\newblock When do neural nets outperform boosted trees on tabular data?
\newblock \emph{arXiv preprint arXiv:2305.02997}, 2023.

\bibitem[M{\"u}ller et~al.(2022)M{\"u}ller, Hollmann, Arango, Grabocka, and Hutter]{muller2022transformers}
Samuel M{\"u}ller, Noah Hollmann, Sebastian~Pineda Arango, Josif Grabocka, and Frank Hutter.
\newblock Transformers can do {Bayesian} inference.
\newblock In \emph{International Conference on Learning Representations}, 2022.

\bibitem[Radford et~al.(2021)Radford, Kim, Hallacy, Ramesh, Goh, Agarwal, Sastry, Askell, Mishkin, Clark, et~al.]{radford2021learning}
Alec Radford, Jong~Wook Kim, Chris Hallacy, Aditya Ramesh, Gabriel Goh, Sandhini Agarwal, Girish Sastry, Amanda Askell, Pamela Mishkin, Jack Clark, et~al.
\newblock Learning transferable visual models from natural language supervision.
\newblock In \emph{International conference on machine learning}, pp.\  8748--8763. PMLR, 2021.

\bibitem[Sattarov et~al.(2023)Sattarov, Schreyer, and Borth]{sattarov2023findiff}
Timur Sattarov, Marco Schreyer, and Damian Borth.
\newblock Findiff: Diffusion models for financial tabular data generation.
\newblock In \emph{Proceedings of the Fourth ACM International Conference on AI in Finance}, pp.\  64--72, 2023.

\bibitem[Somepalli et~al.(2021)Somepalli, Goldblum, Schwarzschild, Bruss, and Goldstein]{somepalli2021saint}
Gowthami Somepalli, Micah Goldblum, Avi Schwarzschild, C~Bayan Bruss, and Tom Goldstein.
\newblock Saint: Improved neural networks for tabular data via row attention and contrastive pre-training.
\newblock \emph{arXiv preprint arXiv:2106.01342}, 2021.

\bibitem[Tang et~al.(2020)Tang, Xia, Zhang, and Long]{tang2020customer}
Qi~Tang, Guoen Xia, Xianquan Zhang, and Feng Long.
\newblock A customer churn prediction model based on {XGBoost and MLP}.
\newblock In \emph{2020 International Conference on Computer Engineering and Application (ICCEA)}, pp.\  608--612, 2020.

\bibitem[Vaswani et~al.(2017)Vaswani, Shazeer, Parmar, Uszkoreit, Jones, Gomez, Kaiser, and Polosukhin]{vaswani2017attention}
Ashish Vaswani, Noam Shazeer, Niki Parmar, Jakob Uszkoreit, Llion Jones, Aidan~N Gomez, {\L}ukasz Kaiser, and Illia Polosukhin.
\newblock Attention is all you need.
\newblock \emph{Advances in neural information processing systems}, 30, 2017.

\bibitem[Wang et~al.(2018)Wang, Zhu, Torralba, and Efros]{wang2018dataset}
Tongzhou Wang, Jun-Yan Zhu, Antonio Torralba, and Alexei~A. Efros.
\newblock Dataset distillation.
\newblock \emph{arXiv preprint arXiv: 1811.10959}, 2018.
\newblock URL \url{https://arxiv.org/abs/1811.10959v3}.

\bibitem[Xu et~al.(2023)Xu, Xie, Qin, Tao, and Wang]{xu2023parameter}
Lingling Xu, Haoran Xie, Si-Zhao~Joe Qin, Xiaohui Tao, and Fu~Lee Wang.
\newblock Parameter-efficient fine-tuning methods for pretrained language models: A critical review and assessment.
\newblock \emph{arXiv preprint arXiv:2312.12148}, 2023.

\bibitem[Zhao et~al.(2021)Zhao, Mopuri, and Bilen]{zhao2021dataset}
Bo~Zhao, Konda~Reddy Mopuri, and Hakan Bilen.
\newblock Dataset condensation with gradient matching.
\newblock In \emph{9th International Conference on Learning Representations, {ICLR} 2021, Virtual Event, Austria, May 3-7, 2021}. OpenReview.net, 2021.
\newblock URL \url{https://openreview.net/forum?id=mSAKhLYLSsl}.

\end{thebibliography}
\bibliographystyle{iclr2024_conference}

\appendix
\section{Appendix}

\subsection{Dataset Details}
\label{section:dataset_details}

We chose 48 large datasets from OpenML \citep{bischl2021openml} to test the generalization performance of TabPFN-ICD against other state-of-the-art methods. The details of these datasets are listed in \autoref{table:openml-datasets}
\begin{table}[h]
    \caption{48 large datasets from OpenML}
    \vspace{5pt}
    \label{table:openml-datasets}
    \centering
    \scriptsize
\begin{tabular}{lrrrrrrr}
\toprule
 & did & \# instances & \# train & \# feat & \# classes & \# cat & class imbalance ratio \\
\midrule
Amazon employee access & 34539 & 32769 & 26215 & 9 & 2 & 9 & 16.258065 \\
Click prediction small & 190408 & 39948 & 31958 & 11 & 2 & 6 & 4.937941 \\
GesturePhaseSegmentationProcessed & 14969 & 9873 & 7897 & 32 & 5 & 0 & 2.957393 \\
JapaneseVowels & 3510 & 9961 & 7968 & 14 & 9 & 0 & 2.070513 \\
MagicTelescope & 3954 & 19020 & 15216 & 10 & 2 & 0 & 1.843049 \\
MiniBooNE & 168335 & 130064 & 104050 & 50 & 2 & 0 & 2.563478 \\
PhishingWebsites & 14952 & 11055 & 8843 & 30 & 2 & 30 & 1.257019 \\
Satellite & 167211 & 5100 & 4080 & 36 & 2 & 0 & 65.885246 \\
ada agnostic & 3896 & 4562 & 3648 & 48 & 2 & 0 & 3.035398 \\
adult-census & 3953 & 32561 & 26048 & 14 & 2 & 8 & 3.153061 \\
adult & 7592 & 48842 & 39072 & 14 & 2 & 8 & 3.179271 \\
artificial-characters & 14964 & 10218 & 8174 & 7 & 10 & 0 & 2.362500 \\
bank-marketing & 14965 & 45211 & 36168 & 16 & 2 & 9 & 7.546314 \\
cardiotocography & 9979 & 2126 & 1700 & 35 & 10 & 0 & 10.767442 \\
churn & 167141 & 5000 & 4000 & 20 & 2 & 4 & 6.054674 \\
connect-4 & 146195 & 67557 & 54045 & 42 & 3 & 42 & 6.896104 \\
eeg-eye-state & 14951 & 14980 & 11984 & 14 & 2 & 0 & 1.227923 \\
electricity & 219 & 45312 & 36248 & 8 & 2 & 1 & 1.355449 \\
elevators & 3711 & 16599 & 13279 & 18 & 2 & 0 & 2.235624 \\
eye movements & 3897 & 10936 & 8748 & 27 & 3 & 3 & 1.484321 \\
first-order-theorem-proving & 9985 & 6118 & 4894 & 51 & 6 & 0 & 5.238462 \\
house 16H & 3686 & 22784 & 18226 & 16 & 2 & 0 & 2.378940 \\
jannis & 168330 & 83733 & 66985 & 54 & 4 & 0 & 22.845070 \\
jungle chess 2pcs raw endgame complete & 167119 & 44819 & 35855 & 6 & 3 & 0 & 5.317959 \\
kc1 & 3917 & 2109 & 1687 & 21 & 2 & 0 & 5.438931 \\
kr-vs-kp & 3 & 3196 & 2556 & 36 & 2 & 36 & 1.093366 \\
magic & 146206 & 19020 & 15216 & 10 & 2 & 0 & 1.843049 \\
mfeat-fourier & 14 & 2000 & 1600 & 76 & 10 & 0 & 1.000000 \\
mfeat-karhunen & 16 & 2000 & 1600 & 64 & 10 & 0 & 1.000000 \\
mfeat-morphological & 18 & 2000 & 1600 & 6 & 10 & 0 & 1.000000 \\
mfeat-zernike & 22 & 2000 & 1600 & 47 & 10 & 0 & 1.000000 \\
mushroom & 24 & 8124 & 6498 & 22 & 2 & 22 & 1.074713 \\
numerai28.6 & 167120 & 96320 & 77056 & 21 & 2 & 0 & 1.020982 \\
nursery & 9892 & 12958 & 10366 & 8 & 4 & 8 & 13.190840 \\
optdigits & 28 & 5620 & 4496 & 64 & 10 & 0 & 1.029279 \\
ozone-level-8hr & 9978 & 2534 & 2026 & 72 & 2 & 0 & 14.828125 \\
page-blocks & 30 & 5473 & 4377 & 10 & 5 & 0 & 178.590909 \\
pendigits & 32 & 10992 & 8792 & 16 & 10 & 0 & 1.086595 \\
phoneme & 9952 & 5404 & 4322 & 5 & 2 & 0 & 2.408517 \\
pollen & 3735 & 3848 & 3078 & 5 & 2 & 0 & 1.001300 \\
satimage & 2074 & 6430 & 5144 & 36 & 6 & 0 & 2.454910 \\
segment & 146822 & 2310 & 1848 & 16 & 7 & 0 & 1.000000 \\
shuttle & 146212 & 58000 & 46400 & 9 & 7 & 0 & 4558.500000 \\
spambase & 43 & 4601 & 3680 & 57 & 2 & 0 & 1.539683 \\
splice & 45 & 3190 & 2552 & 60 & 3 & 60 & 2.161501 \\
sylvine & 168912 & 5124 & 4098 & 20 & 2 & 0 & 1.000977 \\
wall-robot-navigation & 9960 & 5456 & 4364 & 24 & 4 & 0 & 6.729008 \\
wilt & 146820 & 4839 & 3871 & 5 & 2 & 0 & 17.521531 \\
\bottomrule
\end{tabular}
\end{table}

\FloatBarrier

\subsection{Baseline Details}

For the baselines, we leverage the TabZilla GitHub repository, accessible at \url{https://github.com/naszilla/tabzilla}. We also use a training batch size of 128 over 100 epochs for all models.

For XGBoost (Tuned), we tuned the hyperparameters over 5 rounds, which led to the following best hyperparameters: 
$\text{max depth}= 11$, $\alpha=0.00023870306386021943$, $\eta=0.15557408249528354$ and $\lambda=0.6046093787689293$

\FloatBarrier

\subsection{Additional Results}

\begin{table}[h!]
  \caption{AUC per dataset}
  \vspace{5pt}
  \label{table:detailed_performance_table}
  \centering
  \scriptsize 
  \setlength{\tabcolsep}{3pt} 
\begin{tabular}{lrrrrrrrrr}
\toprule
 & RandomForest & TabPFN-ICD & MLP & saint & TabNet & TabPFN & vime & XGBoost & XGBoost (Tuned) \\
dataset &  &  &  &  &  &  &  &  &  \\
\midrule
Amazon-employee-access-34539 & 0.732 & 0.759 & 0.626 & 0.815 & 0.721 & 0.670 & 0.500 & 0.805 & 0.866 \\
Click-prediction-small-190408 & 0.668 & 0.639 & 0.584 & 0.622 & 0.599 & 0.619 & 0.571 & 0.680 & 0.652 \\
GesturePhaseSegmentationProcessed-14969 & 0.788 & 0.845 & 0.711 & 0.750 & 0.778 & 0.775 & 0.657 & 0.864 & 0.902 \\
JapaneseVowels-3510 & 0.980 & 1.000 & 0.970 & 1.000 & 1.000 & 0.997 & 0.993 & 0.999 & 1.000 \\
MagicTelescope-3954 & 0.897 & 0.924 & 0.889 & 0.875 & 0.938 & 0.923 & 0.889 & 0.937 & 0.939 \\
MiniBooNE-168335 & 0.958 & 0.979 & 0.949 & 0.962 & 0.939 & 0.951 & 0.911 & 0.979 & 0.984 \\
PhishingWebsites-14952 & 0.979 & 0.992 & 0.989 & 0.992 & 0.994 & 0.983 & 0.978 & 0.992 & 0.995 \\
Satellite-167211 & 0.993 & 0.993 & 0.988 & 0.771 & 0.908 & 0.991 & 0.827 & 0.994 & 0.988 \\
ada-agnostic-3896 & 0.882 & 0.888 & 0.876 & 0.818 & 0.879 & 0.885 & 0.818 & 0.894 & 0.883 \\
adult-census-3953 & 0.899 & 0.889 & 0.658 & 0.865 & 0.906 & 0.885 & 0.616 & 0.923 & 0.914 \\
adult-7592 & 0.902 & 0.900 & 0.719 & 0.863 & 0.915 & 0.898 & 0.563 & 0.929 & 0.927 \\
artificial-characters-14964 & 0.915 & 0.972 & 0.799 & 0.955 & 0.962 & 0.951 & 0.930 & 0.973 & 0.995 \\
bank-marketing-14965 & 0.900 & 0.906 & 0.885 & 0.807 & 0.933 & 0.889 & 0.849 & 0.927 & 0.927 \\
cardiotocography-9979 & 1.000 & 1.000 & 0.760 & 1.000 & 1.000 & 1.000 & 0.512 & 1.000 & 1.000 \\
churn-167141 & 0.859 & 0.876 & 0.725 & 0.868 & 0.899 & 0.866 & 0.679 & 0.857 & 0.857 \\
connect-4-146195 & 0.717 & 0.875 & 0.752 & 0.873 & 0.878 & 0.643 & 0.502 & 0.832 & 0.909 \\
eeg-eye-state-14951 & 0.856 & 0.990 & 0.595 & 0.477 & 0.685 & 0.969 & 0.611 & 0.953 & 0.987 \\
electricity-219 & 0.864 & 0.937 & 0.888 & 0.878 & 0.900 & 0.854 & 0.822 & 0.933 & 0.981 \\
elevators-3711 & 0.841 & 0.936 & 0.762 & 0.652 & 0.935 & 0.932 & 0.718 & 0.917 & 0.928 \\
eye-movements-3897 & 0.747 & 0.786 & 0.665 & 0.761 & 0.777 & 0.735 & 0.682 & 0.826 & 0.893 \\
first-order-theorem-proving-9985 & 0.754 & 0.781 & 0.641 & 0.783 & 0.717 & 0.739 & 0.519 & 0.806 & 0.810 \\
house-16H-3686 & 0.924 & 0.947 & 0.851 & 0.788 & 0.918 & 0.924 & 0.782 & 0.951 & 0.951 \\
jannis-168330 & 0.771 & 0.819 & 0.774 & 0.816 & 0.849 & 0.771 & 0.739 & 0.840 & 0.848 \\
jungle-chess-2pcs-raw-endgame-complete-167119 & 0.872 & 0.979 & 0.849 & 0.995 & 0.978 & 0.919 & 0.844 & 0.952 & 0.970 \\
kc1-3917 & 0.810 & 0.823 & 0.768 & 0.820 & 0.806 & 0.826 & 0.791 & 0.789 & 0.825 \\
kr-vs-kp-3 & 0.994 & 1.000 & 1.000 & 1.000 & 0.996 & 0.999 & 0.980 & 1.000 & 1.000 \\
magic-146206 & 0.897 & 0.927 & 0.895 & 0.879 & 0.936 & 0.915 & 0.884 & 0.937 & 0.939 \\
mfeat-fourier-14 & 0.975 & 0.976 & 0.945 & 0.989 & 0.973 & 0.976 & 0.930 & 0.983 & 0.980 \\
mfeat-karhunen-16 & 0.992 & 0.998 & 0.998 & 0.997 & 0.998 & 0.993 & 0.999 & 0.997 & 0.997 \\
mfeat-morphological-18 & 0.942 & 0.951 & 0.483 & 0.933 & 0.954 & 0.951 & 0.459 & 0.945 & 0.939 \\
mfeat-zernike-22 & 0.963 & 0.978 & 0.975 & 0.902 & 0.978 & 0.980 & 0.974 & 0.969 & 0.964 \\
mushroom-24 & 1.000 & 1.000 & 1.000 & 1.000 & 1.000 & 1.000 & 1.000 & 1.000 & 1.000 \\
numerai28.6-167120 & 0.525 & 0.526 & 0.526 & 0.508 & 0.511 & 0.515 & 0.521 & 0.522 & 0.508 \\
nursery-9892 & 0.943 & 1.000 & 0.826 & 0.995 & 0.998 & 0.996 & 0.999 & 1.000 & 1.000 \\
optdigits-28 & 0.996 & 0.999 & 1.000 & 1.000 & 0.998 & 0.998 & 0.999 & 1.000 & 1.000 \\
ozone-level-8hr-9978 & 0.933 & 0.945 & 0.931 & 0.938 & 0.914 & 0.945 & 0.832 & 0.910 & 0.920 \\
page-blocks-30 & 0.962 & 0.967 & 0.643 & 0.960 & 0.951 & 0.971 & 0.606 & 0.984 & 0.968 \\
pendigits-32 & 0.996 & 1.000 & 0.965 & 0.997 & 1.000 & 1.000 & 0.999 & 1.000 & 1.000 \\
phoneme-9952 & 0.909 & 0.954 & 0.923 & 0.938 & 0.931 & 0.922 & 0.919 & 0.945 & 0.963 \\
pollen-3735 & 0.489 & 0.462 & 0.496 & 0.548 & 0.499 & 0.462 & 0.504 & 0.507 & 0.480 \\
satimage-2074 & 0.975 & 0.988 & 0.976 & 0.985 & 0.985 & 0.984 & 0.972 & 0.988 & 0.990 \\
segment-146822 & 0.987 & 0.993 & 0.978 & 0.993 & 0.991 & 0.994 & 0.977 & 0.994 & 0.994 \\
shuttle-146212 & 1.000 & 1.000 & 0.557 & 0.859 & 0.977 & 0.999 & 0.462 & 1.000 & 1.000 \\
spambase-43 & 0.963 & 0.978 & 0.971 & 0.967 & 0.971 & 0.971 & 0.954 & 0.983 & 0.984 \\
splice-45 & 0.989 & 0.967 & 0.955 & 0.995 & 0.981 & 0.957 & 0.919 & 0.992 & 0.994 \\
sylvine-168912 & 0.958 & 0.967 & 0.916 & 0.505 & 0.953 & 0.964 & 0.870 & 0.976 & 0.980 \\
wall-robot-navigation-9960 & 0.996 & 0.988 & 0.734 & 0.992 & 0.996 & 0.973 & 0.811 & 1.000 & 1.000 \\
wilt-146820 & 0.982 & 0.994 & 0.893 & 0.661 & 0.993 & 0.996 & 0.710 & 0.990 & 0.989 \\
median & 0.928 & 0.967 & 0.864 & 0.878 & 0.939 & 0.951 & 0.824 & 0.953 & 0.969 \\
\bottomrule
\end{tabular}

\end{table}

\begin{table}[h!]
  \caption{F1 scores per dataset}
  \vspace{5pt}
  \label{table:detailed_performance_table}
  \centering
  \scriptsize 
  \setlength{\tabcolsep}{3pt} 
\begin{tabular}{lrrrrrrrrr}
\toprule
 & RandomForest & TabPFN-ICD & MLP & saint & TabNet & TabPFN & vime & XGBoost & XGBoost (Tuned) \\
dataset &  &  &  &  &  &  &  &  &  \\
\midrule
Amazon-employee-access-34539 & 0.943 & 0.920 & 0.942 & 0.947 & 0.943 & 0.736 & 0.942 & 0.943 & 0.946 \\
Click-prediction-small-190408 & 0.832 & 0.765 & 0.832 & 0.833 & 0.832 & 0.603 & 0.832 & 0.835 & 0.831 \\
GesturePhaseSegmentationProcessed-14969 & 0.425 & 0.640 & 0.410 & 0.432 & 0.464 & 0.463 & 0.332 & 0.602 & 0.678 \\
JapaneseVowels-3510 & 0.821 & 0.994 & 0.823 & 0.977 & 0.983 & 0.932 & 0.918 & 0.963 & 0.980 \\
MagicTelescope-3954 & 0.842 & 0.872 & 0.841 & 0.828 & 0.879 & 0.849 & 0.700 & 0.876 & 0.887 \\
MiniBooNE-168335 & 0.895 & 0.934 & 0.886 & 0.909 & 0.818 & 0.882 & 0.811 & 0.933 & 0.941 \\
PhishingWebsites-14952 & 0.924 & 0.962 & 0.944 & 0.948 & 0.959 & 0.934 & 0.846 & 0.947 & 0.967 \\
Satellite-167211 & 0.990 & 0.986 & 0.990 & 0.984 & 0.990 & 0.990 & 0.986 & 0.990 & 0.990 \\
ada-agnostic-3896 & 0.845 & 0.827 & 0.840 & 0.786 & 0.832 & 0.796 & 0.751 & 0.856 & 0.840 \\
adult-census-3953 & 0.842 & 0.829 & 0.802 & 0.824 & 0.850 & 0.806 & 0.759 & 0.869 & 0.861 \\
adult-7592 & 0.845 & 0.843 & 0.805 & 0.816 & 0.857 & 0.809 & 0.761 & 0.871 & 0.876 \\
artificial-characters-14964 & 0.476 & 0.744 & 0.481 & 0.624 & 0.639 & 0.599 & 0.610 & 0.759 & 0.915 \\
bank-marketing-14965 & 0.892 & 0.892 & 0.892 & 0.848 & 0.910 & 0.845 & 0.883 & 0.906 & 0.906 \\
cardiotocography-9979 & 0.983 & 1.000 & 0.401 & 1.000 & 1.000 & 1.000 & 0.117 & 1.000 & 1.000 \\
churn-167141 & 0.894 & 0.931 & 0.872 & 0.912 & 0.954 & 0.874 & 0.860 & 0.950 & 0.948 \\
connect-4-146195 & 0.530 & 0.810 & 0.746 & 0.805 & 0.813 & 0.521 & 0.523 & 0.729 & 0.841 \\
eeg-eye-state-14951 & 0.778 & 0.954 & 0.578 & 0.551 & 0.632 & 0.909 & 0.545 & 0.879 & 0.944 \\
electricity-219 & 0.774 & 0.862 & 0.807 & 0.803 & 0.819 & 0.772 & 0.602 & 0.853 & 0.931 \\
elevators-3711 & 0.817 & 0.897 & 0.769 & 0.707 & 0.891 & 0.875 & 0.691 & 0.872 & 0.881 \\
eye-movements-3897 & 0.520 & 0.599 & 0.456 & 0.545 & 0.556 & 0.491 & 0.467 & 0.630 & 0.724 \\
first-order-theorem-proving-9985 & 0.429 & 0.542 & 0.472 & 0.571 & 0.437 & 0.477 & 0.247 & 0.567 & 0.586 \\
house-16H-3686 & 0.862 & 0.887 & 0.794 & 0.764 & 0.846 & 0.849 & 0.706 & 0.887 & 0.891 \\
jannis-168330 & 0.630 & 0.681 & 0.668 & 0.681 & 0.703 & 0.550 & 0.637 & 0.694 & 0.714 \\
jungle-chess-2pcs-raw-endgame-complete-167119 & 0.708 & 0.879 & 0.797 & 0.952 & 0.883 & 0.783 & 0.667 & 0.841 & 0.847 \\
kc1-3917 & 0.872 & 0.871 & 0.848 & 0.848 & 0.867 & 0.796 & 0.848 & 0.863 & 0.858 \\
kr-vs-kp-3 & 0.941 & 0.991 & 0.991 & 0.994 & 0.988 & 0.988 & 0.828 & 0.994 & 0.994 \\
magic-146206 & 0.842 & 0.880 & 0.840 & 0.826 & 0.880 & 0.843 & 0.707 & 0.876 & 0.887 \\
mfeat-fourier-14 & 0.745 & 0.793 & 0.663 & 0.858 & 0.790 & 0.752 & 0.584 & 0.861 & 0.837 \\
mfeat-karhunen-16 & 0.925 & 0.925 & 0.955 & 0.954 & 0.945 & 0.895 & 0.970 & 0.933 & 0.944 \\
mfeat-morphological-18 & 0.641 & 0.701 & 0.018 & 0.633 & 0.692 & 0.686 & 0.018 & 0.655 & 0.639 \\
mfeat-zernike-22 & 0.704 & 0.802 & 0.786 & 0.469 & 0.783 & 0.803 & 0.776 & 0.757 & 0.779 \\
mushroom-24 & 0.993 & 1.000 & 1.000 & 1.000 & 1.000 & 1.000 & 0.985 & 1.000 & 1.000 \\
numerai28.6-167120 & 0.516 & 0.515 & 0.518 & 0.504 & 0.513 & 0.510 & 0.512 & 0.519 & 0.506 \\
nursery-9892 & 0.863 & 0.999 & 0.948 & 0.921 & 0.956 & 0.944 & 0.984 & 0.997 & 0.997 \\
optdigits-28 & 0.955 & 0.991 & 0.964 & 0.970 & 0.979 & 0.941 & 0.974 & 0.975 & 0.973 \\
ozone-level-8hr-9978 & 0.945 & 0.939 & 0.937 & 0.937 & 0.945 & 0.922 & 0.937 & 0.937 & 0.941 \\
page-blocks-30 & 0.967 & 0.962 & 0.941 & 0.961 & 0.942 & 0.934 & 0.912 & 0.966 & 0.964 \\
pendigits-32 & 0.904 & 0.997 & 0.942 & 0.942 & 0.990 & 0.988 & 0.986 & 0.989 & 0.989 \\
phoneme-9952 & 0.835 & 0.915 & 0.858 & 0.880 & 0.872 & 0.841 & 0.728 & 0.884 & 0.913 \\
pollen-3735 & 0.499 & 0.439 & 0.514 & 0.504 & 0.494 & 0.439 & 0.506 & 0.506 & 0.499 \\
satimage-2074 & 0.872 & 0.901 & 0.876 & 0.891 & 0.904 & 0.879 & 0.884 & 0.899 & 0.917 \\
segment-146822 & 0.866 & 0.948 & 0.912 & 0.917 & 0.890 & 0.939 & 0.916 & 0.930 & 0.926 \\
shuttle-146212 & 0.998 & 1.000 & 0.914 & 0.791 & 0.997 & 0.971 & 0.692 & 1.000 & 1.000 \\
spambase-43 & 0.896 & 0.939 & 0.924 & 0.926 & 0.937 & 0.909 & 0.764 & 0.933 & 0.937 \\
splice-45 & 0.944 & 0.877 & 0.838 & 0.954 & 0.944 & 0.823 & 0.685 & 0.947 & 0.941 \\
sylvine-168912 & 0.901 & 0.920 & 0.844 & 0.499 & 0.904 & 0.908 & 0.499 & 0.938 & 0.942 \\
wall-robot-navigation-9960 & 0.965 & 0.945 & 0.652 & 0.929 & 0.945 & 0.843 & 0.765 & 0.998 & 0.998 \\
wilt-146820 & 0.969 & 0.980 & 0.948 & 0.946 & 0.983 & 0.977 & 0.946 & 0.979 & 0.979 \\
median & 0.862 & 0.899 & 0.840 & 0.853 & 0.887 & 0.847 & 0.760 & 0.893 & 0.921 \\
\bottomrule
\end{tabular}

\end{table}

\begin{table}[h!]
  \caption{Accuracies per dataset}
  \vspace{5pt}
  \label{table:detailed_performance_table}
  \centering
  \scriptsize 
  \setlength{\tabcolsep}{3pt} 
\begin{tabular}{lrrrrrrrrr}
\toprule
 & RandomForest & TabPFN-ICD & MLP & saint & TabNet & TabPFN & vime & XGBoost & XGBoost (Tuned) \\
dataset &  &  &  &  &  &  &  &  &  \\
\midrule
Amazon-employee-access-34539 & 0.943 & 0.941 & 0.942 & 0.947 & 0.943 & 0.640 & 0.942 & 0.943 & 0.946 \\
Click-prediction-small-190408 & 0.832 & 0.823 & 0.832 & 0.833 & 0.832 & 0.547 & 0.832 & 0.835 & 0.831 \\
GesturePhaseSegmentationProcessed-14969 & 0.501 & 0.644 & 0.481 & 0.470 & 0.486 & 0.459 & 0.454 & 0.616 & 0.689 \\
JapaneseVowels-3510 & 0.832 & 0.994 & 0.855 & 0.977 & 0.983 & 0.932 & 0.918 & 0.963 & 0.980 \\
MagicTelescope-3954 & 0.842 & 0.873 & 0.841 & 0.828 & 0.879 & 0.848 & 0.700 & 0.876 & 0.887 \\
MiniBooNE-168335 & 0.895 & 0.934 & 0.886 & 0.909 & 0.818 & 0.879 & 0.811 & 0.933 & 0.941 \\
PhishingWebsites-14952 & 0.924 & 0.962 & 0.944 & 0.948 & 0.959 & 0.934 & 0.846 & 0.947 & 0.967 \\
Satellite-167211 & 0.990 & 0.988 & 0.990 & 0.984 & 0.990 & 0.988 & 0.986 & 0.990 & 0.990 \\
ada-agnostic-3896 & 0.845 & 0.836 & 0.840 & 0.786 & 0.832 & 0.786 & 0.751 & 0.856 & 0.840 \\
adult-census-3953 & 0.842 & 0.834 & 0.802 & 0.824 & 0.850 & 0.795 & 0.759 & 0.869 & 0.861 \\
adult-7592 & 0.845 & 0.849 & 0.805 & 0.816 & 0.857 & 0.796 & 0.761 & 0.871 & 0.876 \\
artificial-characters-14964 & 0.494 & 0.745 & 0.549 & 0.624 & 0.645 & 0.607 & 0.615 & 0.762 & 0.915 \\
bank-marketing-14965 & 0.892 & 0.900 & 0.892 & 0.848 & 0.910 & 0.819 & 0.883 & 0.906 & 0.906 \\
cardiotocography-9979 & 0.986 & 1.000 & 0.531 & 1.000 & 1.000 & 1.000 & 0.272 & 1.000 & 1.000 \\
churn-167141 & 0.894 & 0.934 & 0.872 & 0.912 & 0.954 & 0.862 & 0.860 & 0.950 & 0.948 \\
connect-4-146195 & 0.661 & 0.824 & 0.777 & 0.815 & 0.837 & 0.481 & 0.658 & 0.774 & 0.854 \\
eeg-eye-state-14951 & 0.778 & 0.954 & 0.578 & 0.551 & 0.632 & 0.909 & 0.545 & 0.879 & 0.944 \\
electricity-219 & 0.774 & 0.862 & 0.807 & 0.803 & 0.819 & 0.771 & 0.602 & 0.853 & 0.931 \\
elevators-3711 & 0.817 & 0.899 & 0.769 & 0.707 & 0.891 & 0.874 & 0.691 & 0.872 & 0.881 \\
eye-movements-3897 & 0.518 & 0.598 & 0.468 & 0.547 & 0.559 & 0.508 & 0.471 & 0.629 & 0.723 \\
first-order-theorem-proving-9985 & 0.503 & 0.569 & 0.526 & 0.590 & 0.495 & 0.453 & 0.418 & 0.593 & 0.596 \\
house-16H-3686 & 0.862 & 0.889 & 0.794 & 0.764 & 0.846 & 0.846 & 0.706 & 0.887 & 0.891 \\
jannis-168330 & 0.644 & 0.685 & 0.675 & 0.685 & 0.711 & 0.533 & 0.644 & 0.701 & 0.720 \\
jungle-chess-2pcs-raw-endgame-complete-167119 & 0.747 & 0.879 & 0.818 & 0.952 & 0.884 & 0.766 & 0.711 & 0.843 & 0.848 \\
kc1-3917 & 0.872 & 0.877 & 0.848 & 0.848 & 0.867 & 0.773 & 0.848 & 0.863 & 0.858 \\
kr-vs-kp-3 & 0.941 & 0.991 & 0.991 & 0.994 & 0.988 & 0.988 & 0.828 & 0.994 & 0.994 \\
magic-146206 & 0.842 & 0.881 & 0.840 & 0.826 & 0.880 & 0.842 & 0.707 & 0.876 & 0.887 \\
mfeat-fourier-14 & 0.750 & 0.795 & 0.705 & 0.860 & 0.790 & 0.765 & 0.620 & 0.860 & 0.840 \\
mfeat-karhunen-16 & 0.925 & 0.925 & 0.955 & 0.955 & 0.945 & 0.895 & 0.970 & 0.935 & 0.945 \\
mfeat-morphological-18 & 0.650 & 0.705 & 0.100 & 0.640 & 0.710 & 0.690 & 0.100 & 0.660 & 0.645 \\
mfeat-zernike-22 & 0.730 & 0.800 & 0.790 & 0.490 & 0.800 & 0.810 & 0.775 & 0.750 & 0.775 \\
mushroom-24 & 0.993 & 1.000 & 1.000 & 1.000 & 1.000 & 1.000 & 0.985 & 1.000 & 1.000 \\
numerai28.6-167120 & 0.516 & 0.518 & 0.518 & 0.504 & 0.513 & 0.511 & 0.512 & 0.519 & 0.506 \\
nursery-9892 & 0.873 & 0.999 & 0.960 & 0.921 & 0.956 & 0.944 & 0.984 & 0.997 & 0.997 \\
optdigits-28 & 0.956 & 0.991 & 0.964 & 0.970 & 0.979 & 0.941 & 0.973 & 0.975 & 0.973 \\
ozone-level-8hr-9978 & 0.945 & 0.945 & 0.937 & 0.937 & 0.945 & 0.909 & 0.937 & 0.937 & 0.941 \\
page-blocks-30 & 0.969 & 0.960 & 0.954 & 0.964 & 0.947 & 0.920 & 0.931 & 0.965 & 0.964 \\
pendigits-32 & 0.905 & 0.997 & 0.945 & 0.942 & 0.990 & 0.988 & 0.985 & 0.989 & 0.989 \\
phoneme-9952 & 0.835 & 0.915 & 0.858 & 0.880 & 0.872 & 0.835 & 0.728 & 0.884 & 0.913 \\
pollen-3735 & 0.499 & 0.457 & 0.514 & 0.504 & 0.494 & 0.457 & 0.506 & 0.506 & 0.499 \\
satimage-2074 & 0.879 & 0.904 & 0.880 & 0.893 & 0.904 & 0.877 & 0.890 & 0.902 & 0.919 \\
segment-146822 & 0.870 & 0.948 & 0.918 & 0.918 & 0.892 & 0.939 & 0.918 & 0.931 & 0.926 \\
shuttle-146212 & 0.998 & 1.000 & 0.939 & 0.851 & 0.998 & 0.953 & 0.786 & 1.000 & 1.000 \\
spambase-43 & 0.896 & 0.939 & 0.924 & 0.926 & 0.937 & 0.909 & 0.764 & 0.933 & 0.937 \\
splice-45 & 0.944 & 0.878 & 0.837 & 0.953 & 0.944 & 0.821 & 0.708 & 0.947 & 0.940 \\
sylvine-168912 & 0.901 & 0.920 & 0.844 & 0.499 & 0.904 & 0.908 & 0.499 & 0.938 & 0.942 \\
wall-robot-navigation-9960 & 0.965 & 0.945 & 0.736 & 0.929 & 0.945 & 0.842 & 0.793 & 0.998 & 0.998 \\
wilt-146820 & 0.969 & 0.979 & 0.948 & 0.946 & 0.983 & 0.975 & 0.946 & 0.979 & 0.979 \\
median & 0.866 & 0.902 & 0.841 & 0.855 & 0.887 & 0.844 & 0.762 & 0.894 & 0.923 \\
\bottomrule
\end{tabular}

\end{table}

\end{document}